\pdfoutput=1

\documentclass[11pt]{article}

\usepackage[]{ACL2023}

\usepackage{times}
\usepackage{latexsym}
\usepackage{amsmath,amssymb,amsfonts}
\usepackage{algorithmic}
\usepackage{graphicx}
\usepackage{textcomp}
\usepackage{xcolor}
\usepackage{times}
\usepackage{latexsym}
\usepackage{cleveref}
\usepackage{supertabular,booktabs}
\usepackage{multirow,makecell}

\usepackage{longtable}

\usepackage[T1]{fontenc}

\usepackage[utf8]{inputenc}

\usepackage{microtype}

\usepackage{inconsolata}

\usepackage[normalem]{ulem} 
\usepackage{xcolor}

\def\footurl#1{\footnote{\url{#1}}}

\usepackage{pgfplots,pgfplotstable}
\pgfplotsset{compat=1.7}
\usepackage[T4,T1]{fontenc}
\usepackage[verbose]{newunicodechar}
\newenvironment{tfour}{\fontencoding{T4}\selectfont}{}

\newcommand\myfontsize{\fontsize{10.5pt}{13.6pt}\selectfont}

\title{HaVQA: A Dataset for Visual Question Answering and Multimodal Research in Hausa Language}

\author{\normalsize Shantipriya Parida$^{1}$, Idris Abdulmumin$^{2,4}$, Shamsuddeen Hassan Muhammad$^{3,4}$, \\
\textbf{\normalsize Aneesh Bose$^{5,10}$, Guneet Singh Kohli$^{6,10}$, Ibrahim Said Ahmad$^{3,4}$, Ketan Kotwal$^7$,}\\
\textbf{\normalsize Sayan Deb Sarkar$^{8,10}$, Ondřej Bojar$^{9}$, Habeebah Adamu Kakudi$^3$}\\[2mm]
\normalsize $^1$Silo AI, Finland, $^2$Ahmadu Bello University, Zaria, Nigeria, $^3$Bayero University Kano, Nigeria,\\
\normalsize $^4$HausaNLP, $^5$Microsoft, India, $^6$Thapar University, India, $^7$Idiap Research Institute, Switzerland,\\ \normalsize $^8$ETH Zurich, Switzerland, $^9$Charles University, MFF \'{U}FAL, Czech Republic, \normalsize $^{10}$CORD.ai, India\\[0.5mm]
\normalsize  
\texttt{correspondence: shantipriya.parida@silo.ai, iabdulmumin@abu.edu.ng
}}

\begin{document}
\maketitle
\begin{abstract}
This paper presents HaVQA, the first multimodal dataset for visual question-answering (VQA) tasks in the Hausa language. 
The dataset was created by manually translating 6,022 English question-answer pairs, which are associated with 1,555 unique images from the Visual Genome dataset. As a result, the dataset provides 12,044 gold standard English-Hausa parallel sentences that were translated in a fashion that guarantees their semantic match with the corresponding visual information. We conducted several baseline experiments on the dataset, including visual question answering, visual question elicitation, text-only and multimodal machine translation.

\end{abstract}

\section{Introduction}
\label{sect_intro}

In recent years, multidisciplinary research, including Computer Vision (CV) and Natural Language Processing (NLP), has attracted many researchers to the tasks of image captioning, cross-modal retrieval, visual common-sense reasoning, and visual question answering (VQA, \citealp{Antol2015,goyal2017making}). In the VQA task, given an image and a natural language question related to the image, the objective is to produce a correct natural language answer as output \cite{kafle2017visual,ren2015exploring}. VQA is one of the challenging tasks in NLP that requires a fine-grained semantic processing of both the image and the question, together with visual reasoning for an accurate answer prediction \cite{yu2019deep}. The general approaches followed by existing VQA models include \textit{i)} extracting features from the questions and the images, and \textit{ii)} utilizing the features to understand the image content to infer the answers.

Recently, research on improving visual question-answering systems using multimodal architectures and sentence embeddings \cite{kodali2022Recent, urooj2020mmft,gupta2020unified,pfeiffer2021xgqa} has seen tremendous growth. However, most of the VQA datasets used in the VQA research are limited to the English language \cite{kafle2017visual}. Although the accuracy of the VQA systems for English improved significantly with the advent of Transformer-based models (e.g., BERT, \citealp{devlin2018bert}), the lack of VQA datasets  has restricted the development of such systems in most languages, especially the low-resource ones \cite{kumar2022mucot}.

The availability of original datasets for state-of-the-art natural language processing tasks has since been appreciated, especially on the African continent. While some of the efforts to create such datasets for African languages are supported by funding such as Facebook's Translation Support for African Languages, the Lacuna Fund\footurl{https://lacunafund.org/}, and many others, including the dataset in this work are driven by the enthusiasm for developing quality NLP solutions for African languages that are useful to the local communities.
\vspace{-1mm}
\paragraph{Contributions:} 

The main contribution of this work is building a multimodal dataset (HaVQA) for the Hausa language, consisting of question-answer pairs along with the associated images and is suitable for many NLP tasks. As per our knowledge, HaVQA is the first VQA dataset for Hausa language, and will enrich Hausa natural language processing (NLP) resources, allowing researchers to conduct VQA and multimodal research in Hausa. 

\section{Related work}
\label{sect_related_lit}
\subsection{Datasets for African NLP}

African languages are low-resourced; many do not have any datasets for everyday NLP tasks. While some datasets exist for some African languages, they are often limited in scope or lack the necessary quality \cite{kreutzer-etal-2022-quality}. For Visual Question Answering, no publicly available dataset exists in any African language. Luckily, there has been a recent surge in efforts by researchers to create datasets for African languages. In this section, we provide an overview of some of the most recent examples of such efforts.

\citet{abdulmumin-etal-2022-hausa} created the multimodal Hausa Visual Genome dataset for machine translation and image captioning. \citet{adelani-etal-2022-thousand} created the MAFAND-MT\footurl{https://github.com/masakhane-io/lafand-mt/tree/main/data/text_files} collection of parallel datasets between 16 African languages and English or French. The HornMT\footurl{https://github.com/asmelashteka/HornMT} dataset was created for machine translation in languages in the Horn of Africa. \citet{akera2022machine} created about 25,000 parallel sentences between 5 Ugandan languages and English, covering topics such as agriculture, health and society.

\citet{muhammad-etal-2022-naijasenti} classified about 30,000 tweets in each of the four major Nigerian languages as either positive, negative, neutral, mixed or indeterminate for sentiment analysis task. Subsequently, the dataset was expanded to include 14 African languages, resulting in the largest sentiment dataset for African languages \cite{muhammad2023afrisenti,muhammad-etal-2023-semeval}. \citet{aliyu2022herdphobia} collected about 4,500 partially code-switched tweets for detecting hate against the Fulani herdsmen in Nigeria.

\citet{adelani-masakhaner2-2022} created MasakhaNER 2.0, the most extensive corpus for named-entity recognition tasks. \citet{https://doi.org/10.48550/arxiv.2208.12081} created the multipurpose Kencorpus, a speech and text corpora for machine translation, text-to-speech, question answering, and part-of-speech tagging tasks for three Kenyan languages.
KenPOS \cite{DVN/KLCKL5_2022} corpus was created for part-of-speech tagging for Kenyan languages. KenSpeech \cite{DVN/YHXJSU_2022} is a transcription of Swahili speech created for text-to-speech tasks.

\begin{figure*}[!htb]
\begin{center}
\includegraphics[width=0.8\textwidth]{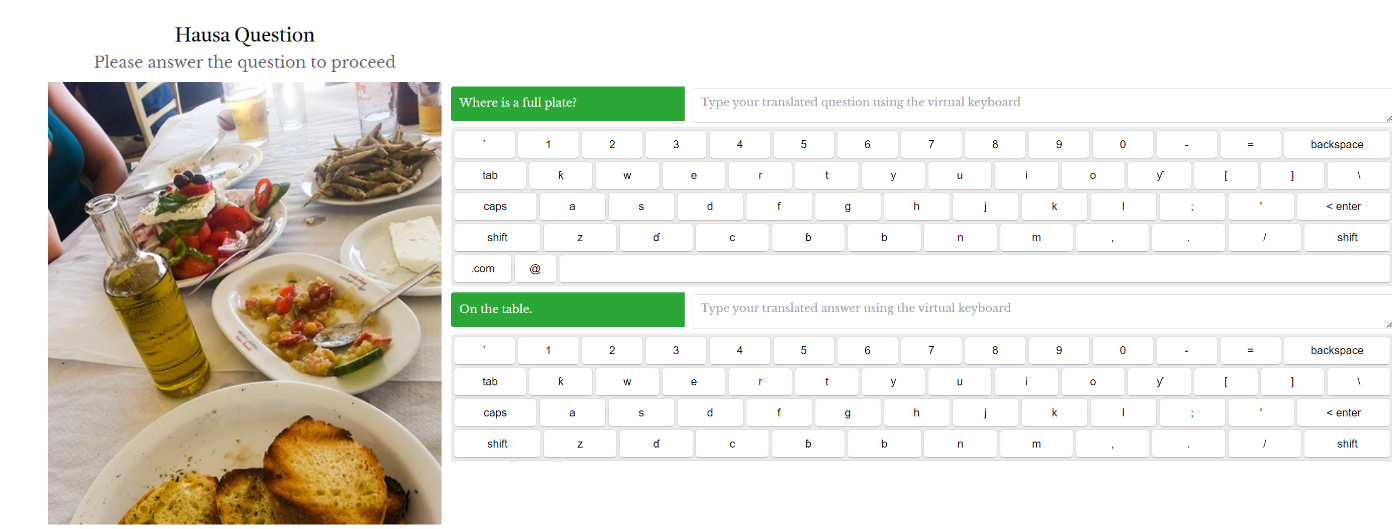}
\caption{Hausa Visual Question Answering (HaVQA) Annotation Interface}
\label{fig_havqa_interface}
\end{center}
\end{figure*}

\subsection{Visual Question Answering Datasets}
Researchers have created several visual question-answering datasets for different purposes, including for medical research \cite{He2021, Lau2018, Abacha2021}, improving reading comprehension \cite{Li2019, Sharma2022}, among others. 

DAQUAR \cite{Malinowski2014} was the first attempt at creating and benchmarking a dataset for understanding and developing models in visual question answering. The dataset consists of 1,449 real-world images and 12,468 synthetic and natural question-answer pairs. \citet{Ren2015} then created a substantially larger dataset called the COCO-QA dataset using images from the Microsoft COCO dataset \cite{Lin2014}. The dataset consists of 123,287 images with each having a corresponding question-answer pair.

\citet{Gao2015} reused the images from COCO-QA to create the FM-IQA dataset. The authors instruct the annotators to create custom questions and answers, through a crowd-sourcing platform. This resulted in 250,560 question-answer pairs from 120,360 images. Other similar datasets include the Visual Madlibs \cite{Yu2015}, VQA \cite{Antol2015}, Visual7W \cite{Zhu2016}, Visual Genome \cite{Krishna2016}, CLEVR \cite{Johnson2017}. Others, such as VQA-HAT \cite{das-etal-2016-human} and VQA-CP \cite{Agrawal_2018_CVPR} datasets, extended the VQA to solve specific problems associated with the original dataset.

In the medical area, \citet{Abacha2021} and \citet{Lau2018} created the VQA-Med and VQA-Rad datasets for radiological research, respectively. The VQA-Med dataset was created from 4,200 radiology images and has multiple-choice 15,292 question-answer pairs. The VQA-Rad data consists of 3,515 questions of 11 types that were crafted manually, where clinicians on radiological images gave both questions and answers. \citet{He2021} created the PathVQA, a datasets that consists of 32,795 mostly open-ended questions and their answer pairs that were generated from 4,998 pathology images.

One common theme across all these datasets is that they are all in English. For the majority of other languages, there exists only a few or no datasets for visual question answering tasks. Some of the few in other languages include the original FM-IQA which was created in Chinese before being manually translated into English. Another is the Japanese VQA \cite{shimizu-etal-2018-visual}, where the authors used crowdsourcing to generate. It consists of 99,208 images, each with eight questions, resulting in 793,664 question-answer pairs in Japanese. For African languages, however, there is no dataset for visual question answering tasks. This is the gap that we are trying to fill with the creation of the HaVQA dataset.

\section{Focused Language}
\label{sect_focus_language}

\textbf{Hausa} is a Chadic language and is the largest indigenous African language that is spoken as the first or second language by about 79 million people, mainly in northern Nigeria, Niger and northern Cameroon, but also in Benin, Burkina Faso, Central African Republic, Chad, Congo, Côte d’Ivoire, Gabon, Sudan and Togo.\footurl{https://www.ethnologue.com/language/hau} The language has well-documented literature and is studied at various local and international institutions. Several international radio stations, including BBC,\footurl{https://www.bbc.com/hausa} VoA,\footurl{https://www.voahausa.com/} and DW\footurl{https://www.dw.com/ha/labarai/s-11605} run Hausa broadcasting service. 

In the early days, the Hausa language was written in ``\textbf{Ajami}'', using Arabic scripts, mainly because of the earlier contacts between the Arabs and the Hausa people \cite{Jaggar2006}. Nowadays, the language is predominantly written in Latin script. This resulted from the colonial influence that began in the early 19th century. Hausa text is written using the English alphabet except for p, q, v and x, with some additional special letters: \begin{tfour}\m{b}\end{tfour}, \begin{tfour}\m{d}\end{tfour}, \begin{tfour}\m{k}\end{tfour} and \begin{tfour}\m{y}\end{tfour}. Nowadays, especially on social media, writing in Hausa has experienced some form of distortion, such as the use of characters p for f, v for b, q or k for \begin{tfour}\m{k}\end{tfour} and d for \begin{tfour}\m{d}\end{tfour}.

\section{The HaVQA Dataset}
\label{sect_dataset}
This section provides a detailed description and analysis of HaVQA Questions and Answers.

\subsection{Data Collection and Annotation}
We extracted the images along with the question-answer (QA) pairs\footurl{https://visualgenome.org/static/data/dataset/question_answers.json.zip} from the Visual Genome dataset \cite{krishna2017visual}. The Visual Genome dataset was created to provide a link between images and natural text, supplying multimodal context in many natural language processing tasks.

We used 7 Hausa native speakers to generate the translations of the QA pair manually. The annotation process was done using a web application developed by integrating the Hausa keyboard to provide easy access to Hausa special characters and restrict access to the unused characters, as shown in \Cref{fig_havqa_interface}. During the translation exercise, a set of instructions was provided to annotators, including (i) ``the translation should be manually generated without using any translation tool'' and (ii) ``the on-screen keyboard or any other keyboard that supports the Hausa special characters should be used.'' These simple and easy-to-remember instructions were adopted to ensure data authenticity and the quality of the dataset. Importantly, the picture has always been presented to the translators.

\subsection{Data Validation}
\label{subsect_dataset_validation}
After the annotation, each question and the answer were validated to ensure the quality and consistency of the translations, including the basic check that the translations were done using the correct alphabet and special characters. We validated the translations by relying on 7 Hausa language experts. A separate interface was created for the validators to be able to see the images, the original English Question-Answer pair, and the translations of all the pairs that were generated by the annotators in the first phase at once.

The common problem was that the annotators mixed up the choice of words when translating objects that did not have a clear masculine or feminine grammatical gender.
Examples of such cases are: ``Ina \begin{tfour}\m{y}\end{tfour}ar tsana ruwan hoda?'' (\textbf{gloss:} Where is the pink teddy bear?), where ``\begin{tfour}\m{y}\end{tfour}ar'' (feminine) was used, but ``yake'' (masculine) was used in ``Me teddy bear \begin{tfour}\m{d}\end{tfour}in yake sanye dashi?'' (\textbf{gloss:} What is the teddy bear wearing?).

Another problem was that the annotators were still using b, d, k and y instead of the special characters \begin{tfour}\m{b}\end{tfour}, \begin{tfour}\m{d}\end{tfour}, \begin{tfour}\m{k}\end{tfour} and \begin{tfour}\m{y}\end{tfour}. Using plain ASCII instead of accented symbols introduces ambiguities that can be sometimes resolved only by consulting the original English questions or answers or the associated image. An example is the different meanings of the words ``kare'' (dog) and ``\begin{tfour}\m{k}\end{tfour}are'' (finish). Some annotations included 'y for \begin{tfour}\m{y}\end{tfour} (e.g., ``Ina 'yar tsanar dabbar?'', instead of ``Ina \begin{tfour}\m{y}\end{tfour}ar tsanar dabbar?'').

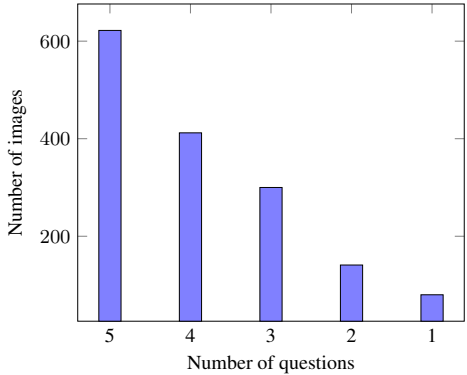
\begin{figure}[t]
    \centering
    \small
    \resizebox{0.8\columnwidth}{!}{%
    \begin{tikzpicture}
        \begin{axis}[
            symbolic x coords={5, 4, 3, 2, 1},
            xtick=data,
            width=\columnwidth,
            ylabel=Number of images,
            xlabel=Number of questions
          ]
            \addplot[ybar,fill=blue!50] coordinates {
                (5, 622)
                (4, 412)
                (3, 300)
                (2, 141)
                (1, 80)
            };
        \end{axis}
    \end{tikzpicture}
    }
    \caption{Number of questions per image}
    \label{fig:image_quest_count}
\end{figure}

\begin{table}[t]
\begin{center}
\footnotesize
\begin{tabular}{ll}
\toprule
\bf Item & \bf Count \\
\midrule
Number of Images & 1,555 \\
Number of Questions & 6,020 \\
Number of Answers & 6,020 \\
Number of Counting Questions & 616 \\
\bottomrule
\end{tabular}
\end{center}
\caption{\label{havqa_dataset_stat} HaVQA Dataset Statistics}
\end{table}
\begin{table}[t]
\begin{center}
\resizebox{\columnwidth}{!}{
\begin{tabular}{llr}
\toprule
\multicolumn{2}{c}{\bf Question Type} & \multirow{2}{*}{\bf \%} \\ \cmidrule{1-2}
\bf Hausa & \bf Gloss & \\
\midrule
``Mene ne/Mene/Me/Wa\begin{tfour}\m{d}\end{tfour}anne'' & What & 56.3 \\
``Me/Mai yasa'' & Why & 2.9 \\
``Yaya (Nawa/Guda nawa)'' & How (much/many) & 12.9 \\
``Yaushe'' & When & 5.6 \\
``A ina/Ina'' & Where & 16.4 \\
``Waye/(wace/wacce/wace ce)'' & Who/(whose) & 5.9 \\
\bottomrule
\end{tabular}
}
\end{center}
\caption{\label{havqa_dataset_question_type} HaVQA Question Types Statistics}
\end{table}

\subsection{Data Analysis}
\label{subsect_dataset_analysis}
The HaVQA consists of questions and their corresponding answers. For each image, at least one and at most five questions were asked and answered; see the distribution in \Cref{fig:image_quest_count}. We used the punkt\footurl{https://www.nltk.org/api/nltk.tokenize.punkt.html} tokenizer in the NLTK toolkit \cite{bird2009natural} for tokenization. Some relevant statistics in the created HaVQA dataset are shown in \Cref{havqa_dataset_stat}.

\subsubsection{Questions} A variety of question types were included in the original English data; they start with the question words: what, why, how, when, where, and who. In the created HaVQA, these words were translated based on the context in which they appeared. In the Hausa language, these question types vary according to usage, gender, and dialect. For example, the word ``who'' is translated as ``\textbf{wanne}'' if it is associated with the male gender, or ``\textbf{wacce}'' or ``\textbf{wace}'' for female. The statistics of the different question types (based on the words that start the question) are shown in \Cref{havqa_dataset_question_type}. The Hausa questions vary, from as short as 2 to as long as 16 words. The length distribution is shown in \Cref{fig:length_precent}.

\begin{figure}
\pgfplotsset{width=9cm,compat=1.14}
\begin{tikzpicture}
\begin{axis}[
        ybar stacked,
        bar width=20pt,
        point meta=explicit symbolic,
        nodes near coords,
        nodes near coords style={
              font=\tiny,
        },
        xlabel={English question words},
        axis lines=left,  enlarge x limits=0.15,
        axis line style={-},
        label style={font=\small},
        ytick={0,1204,2408,3612},
        yticklabel={\pgfmathparse{\tick/6020*100}\pgfmathprintnumber{\pgfmathresult}\%},
        symbolic x coords={what, where, how (1st), how (2nd), who, when, why},
        xtick=data,
        x tick label style={rotate=18,anchor=north,font=\small},
        y tick label style={rotate=90,font=\small},
        height=10cm,
        width=8.5cm
    ]
    \addplot+[ybar] plot coordinates {(what,1134) [menene] (where,0) (how (1st),0) (how (2nd),0) (who,0) (when,0) (why,0)};
    \addplot+[ybar] plot coordinates {(what,542) [me] (where,0) (how (1st),0) (how (2nd),0) (who,0) (when,0) (why,0)};
    \addplot+[ybar] plot coordinates {(what,524) [mene] (where,0) (how (1st),0) (how (2nd),0) (who,0) (when,0) (why,0)};
    \addplot+[ybar] plot coordinates {(what,382) [wani] (where,0) (how (1st),0) (how (2nd),0) (who,0) (when,0) (why,0)};
    \addplot+[ybar] plot coordinates {(what,151) [wanne] (where,0) (how (1st),0) (how (2nd),0) (who,0) (when,0) (why,0)};
    \addplot+[ybar] plot coordinates {(what,131) [others] (where,0) (how (1st),0) (how (2nd),0) (who,0) (when,0) (why,0)};
    \addplot+[ybar] plot coordinates {(what,96) [wacce] (where,0) (how (1st),0) (how (2nd),0) (who,0) (when,0) (why,0)};
    \addplot+[ybar] plot coordinates {(what,77) [wana] (where,0) (how (1st),0) (how (2nd),0) (who,0) (when,0) (why,0)};
    \addplot+[ybar] plot coordinates {(what,61) [a] (where,0) (how (1st),0) (how (2nd),0) (who,0) (when,0) (why,0)};
    \addplot+[ybar] plot coordinates {(what,289) [others] (where,0) (how (1st),0) (how (2nd),0) (who,0) (when,0) (why,0)};
    \addplot+[ybar] plot coordinates {(what,0) (where,637) [ina] (how (1st),0) (how (2nd),0) (who,0) (when,0) (why,0)};
    \addplot+[ybar] plot coordinates {(what,0) (where,332) [a ina] (how (1st),0) (how (2nd),0) (who,0) (when,0) (why,0)};
    \addplot+[ybar] plot coordinates {(what,0) (where,16) [others] (how (1st),0) (how (2nd),0) (who,0) (when,0) (why,0)};
    \addplot+[ybar] plot coordinates {(what,0) (where,0) (how (1st),384) [others] (how (2nd),0) (who,0) (when,0) (why,0)};
    \addplot+[ybar] plot coordinates {(what,0) (where,0) (how (1st),133) [nawane] (how (2nd),0) (who,0) (when,0) (why,0)};
    \addplot+[ybar] plot coordinates {(what,0) (where,0) (how (1st),98) [yaya] (how (2nd),0) (who,0) (when,0) (why,0)};
    \addplot+[ybar] plot coordinates {(what,0) (where,0) (how (1st),95) [mutane] (how (2nd),0) (who,0) (when,0) (why,0)};
    \addplot+[ybar] plot coordinates {(what,0) (where,0) (how (1st),68) [nawa ne] (how (2nd),0) (who,0) (when,0) (why,0)};
    \addplot+[ybar] plot coordinates {(what,0) (where,0) (how (1st),0) (how (2nd),271) [nawa] (who,0) (when,0) (why,0)};
    \addplot+[ybar] plot coordinates {(what,0) (where,0) (how (1st),0) (how (2nd),126) [adadin] (who,0) (when,0) (why,0)};
    \addplot+[ybar] plot coordinates {(what,0) (where,0) (how (1st),0) (how (2nd),0) (who,205) [waye] (when,0) (why,0)};
    \addplot+[ybar] plot coordinates {(what,0) (where,0) (how (1st),0) (how (2nd),0) (who,45) [wa] (when,0) (why,0)};
    \addplot+[ybar] plot coordinates {(what,0) (where,0) (how (1st),0) (how (2nd),0) (who,41) [wanene] (when,0) (why,0)};
    \addplot+[ybar] plot coordinates {(what,0) (where,0) (how (1st),0) (how (2nd),0) (who,33) [wake] (when,0) (why,0)};
    \addplot+[ybar] plot coordinates {(what,0) (where,0) (how (1st),0) (how (2nd),0) (who,33) [others] (when,0) (why,0)};
    \addplot+[ybar] plot coordinates {(what,0) (where,0) (how (1st),0) (how (2nd),0) (who,0) (when,202) [yaushe] (why,0)};
    \addplot+[ybar] plot coordinates {(what,0) (where,0) (how (1st),0) (how (2nd),0) (who,0) (when,80) [da yaushe] (why,0)};
    \addplot+[ybar] plot coordinates {(what,0) (where,0) (how (1st),0) (how (2nd),0) (who,0) (when,55) [others] (why,0)};
    \addplot+[ybar] plot coordinates {(what,0) (where,0) (how (1st),0) (how (2nd),0) (who,0) (when,0) (why,168) [meyasa]};
\end{axis}
\end{tikzpicture}
\caption{Distribution of Hausa words used in asking questions in relation to their English counterparts. We provide the gloss of these words in \Cref{tab:glossaries}.}
    \label{fig:hau_eng_q_tags}
\end{figure}
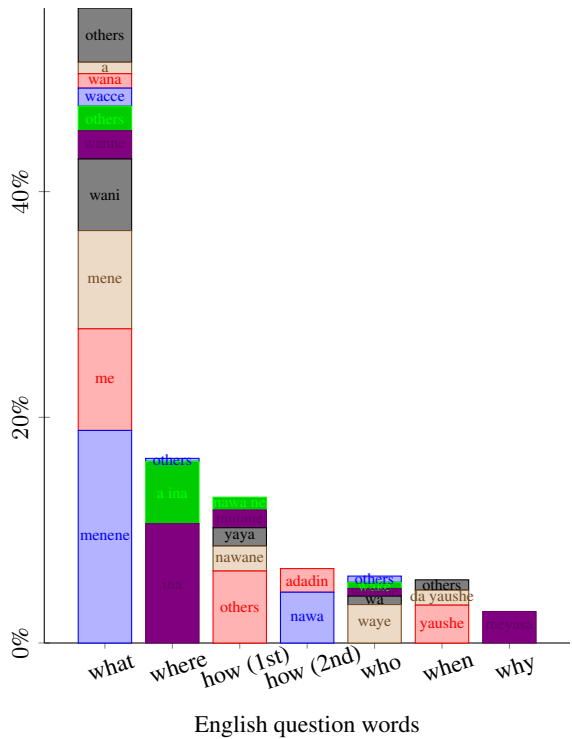

In \Cref{fig:hau_eng_q_tags}, we show the distribution of Hausa words used when asking a question based on the English question tags of the original dataset. While most of the question tags are used at the beginning of the sentence, as in English, ``nawa/nawane'' (how much) is mostly used when the subjects that need counting are mentioned, e.g., 2$^{nd}$ in the example ``Jirgin leda \textbf{nawa} ne a jikin wannan hoton?'' (How many kites are there in this picture?) and 3$^{rd}$ in the example ``Maza \textbf{nawa} ne akan dusar \begin{tfour}\m{k}\end{tfour}an\begin{tfour}\m{k}\end{tfour}arar?'' (How many men are there in the snow?).

\pgfplotstableread[row sep=\\,col sep=&]{
    2 & 5.166112956810631\\
    3 & 17.491694352159467\\
    4 & 26.029900332225914\\
    5 & 23.770764119601328\\
    6 & 13.471760797342192\\
    7 & 6.893687707641195\\
    8 & 3.6212624584717608\\
    9 & 1.8272425249169437\\
    10 & 0.9468438538205981\\
    11 & 0.46511627906976744\\
    12 & 0.16611295681063123\\
    13 & 0.08305647840531562\\
    14 & 0.016611295681063124\\
    15 & 0.03322259136212625\\
    16 & 0.016611295681063124\\
}\dataLengthPercentQuestions

\pgfplotstableread[row sep=\\,col sep=&]{
    1 & 47.99003322259136\\
    2 & 28.122923588039868\\
    3 & 13.704318936877078\\
    4 & 6.544850498338871\\
    5 & 2.1096345514950166\\
    6 & 0.8803986710963454\\
    7 & 0.3986710963455149\\
    8 & 0.11627906976744186\\
    9 & 0.09966777408637872\\
    10 & 0.016611295681063124\\
    11 & 0.016611295681063124\\
}\dataLengthPercentAnswers

\begin{figure}[t]
\centering
\small
    \begin{tikzpicture}
    \begin{axis}[
        xlabel={\# of words in sentences},
        axis lines=left,
        xmin=0.5, xmax=16.5,
        ymin=-2, ymax=52,
        xtick={0,1,2,4,6,8,10,12,14,16},
        ytick={0,10,20,30,40,50},
        yticklabel={\pgfmathprintnumber\tick\%},
        width = \columnwidth,
        height = 6cm
    ]
     
    \addplot[
        color=red,
        mark=o,
        ]
        table {\dataLengthPercentQuestions};
        \addlegendentry{\emph{questions}}
    
     \addplot[
        color=black,
        mark=triangle,
        ]
        table {\dataLengthPercentAnswers};
        \addlegendentry{\emph{answers}}
    
    \end{axis}
    \end{tikzpicture}
    \caption{Percentage of Hausa questions and answers with different word counts in HaVQA.}
    \label{fig:length_precent}
\end{figure}
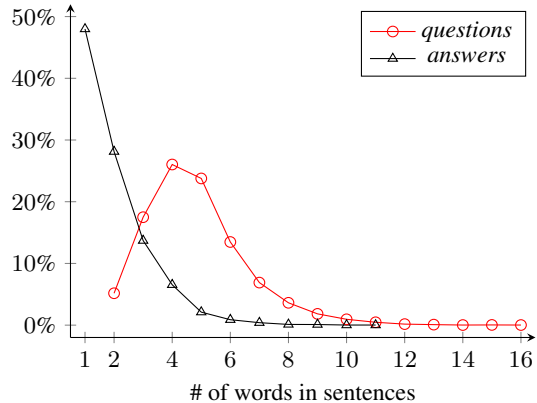

\subsubsection{Answers} Based on the questions, various answers are included, ranging from a single word (or number) to a short description. The distribution of answer lengths is shown in \Cref{fig:length_precent}. The distribution of answers for the question types is shown in \Cref{fig:hau_ques_ans_dist}.

\begin{figure*}[!t]
    \centering
    \includegraphics[clip, trim=2.8cm 13cm 2.3cm 4.2cm, width=\textwidth]{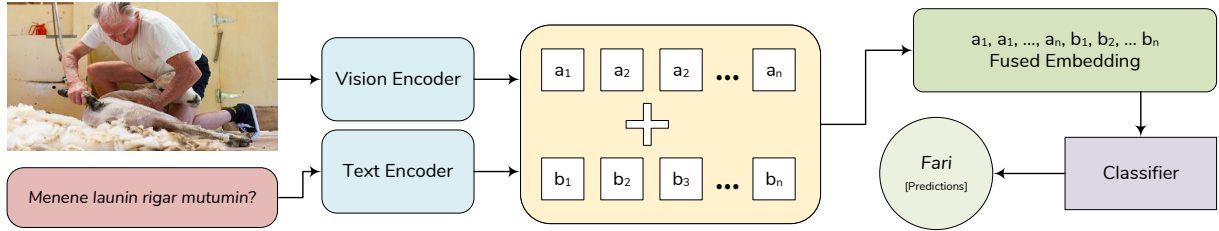}
    \caption{Visual Question Answering. The question is encoded by Hausa BERT, and the context [image] is encoded using a Vision Transformer. The created fused embedding is passed through a classifier to yield the best possible answer. \textbf{Gloss:} Input--What color is the man's shirt?; Prediction--White
    }
    \label{fig:vqa_hausa}
\end{figure*}

\section{Sample Applications of HaVQA}
\label{sect_application}
We tested the HaVQA data by experimenting with the following NLP tasks: \textit{i)} Questions, Answers, and Images for visual question answering, multi-modal machine translation,  \textit{ii)} Questions, and Images for visual question elicitation, and \textit{iii)} Questions and Answers for text-only machine translation.

\subsection{Visual Question Answering}
\label{sect_vqa}
We used multimodal Transformer-based architecture for VQA consisting of three modules: \textit{i)} feature extraction module---which extracts features from the image and question, \textit{ii)} fusion module---which combines both textual and image features, and \textit{iii)} classification module---which obtains the answer \cite{siebert2022multi}.

Similarly, we used Visual Transformer (ViT) for image feature extraction \cite{dosoViTskiy2020image} and multilingual BERT for Hausa, i.e., Hausa BERT\footurl{https://huggingface.co/Davlan/bert-base-multilingual-cased-finetuned-hausa} for extracting features from the Hausa questions. The classifier which obtains the answer is a fully connected network with output having dimensions equal to the answer space. This architecture is illustrated in \Cref{fig:vqa_hausa}. We used the Wu and Palmer metric for VQA evaluation \cite{wu1994verbs}.
\begin{table}[t]
    \centering
    \footnotesize
    \begin{tabular}{lrrr}
    \toprule
        \textbf{Set} & \textbf{Q/A pairs} & \textbf{Tokens (En)}
        & \textbf{Tokens (Ha)}\\
        \midrule
        Train   & 4816   &35,634 &32,142 \\
        Dev     & 602   &4,508 &4,112 \\
        Test    & 602   &4,554 &4,084 \\
        \midrule
        Total   & 6,020  &44,696 & 40338 \\
        \bottomrule
    \end{tabular}
    \caption{Statistics of our data (questions) used in the \textbf{\textit{Visual Question Answering}} task: the number of sentences and tokens.}
    \label{tab:train_test_ie}
\end{table}

\subsection{Machine Translation}
\label{sect_mt}
We performed text-only and multi-modal translation using the HaVQA dataset. We partitioned the dataset into train/dev/test sets in the ratio of 80:10:10 as shown in \Cref{tab:train_test}.

\subsubsection{Text-Only Translation}
\label{sect_mt_text}

We used the questions and answers in English and Hausa for text-only translation. We utilized two approaches for training the Transformer model \cite{t2t_vaswani:et:al}: training from scratch and fine-tuning a pre-trained multi-lingual model. We evaluated the models' performance using SacreBLEU \cite{post-2018-call} for the dev and test set.

\paragraph{Transformer Trained from Scratch}
We used the Transformer model as implemented in OpenNMT-py
\cite{opennmt}.\footurl{http://opennmt.net/OpenNMT-py/quickstart.html} Subword units were constructed using the word pieces algorithm \cite{google_word_pieces:melvin:etal}. Tokenization is handled automatically as part of the pre-processing pipeline of word pieces.

We jointly generated a vocabulary of 32k subword types for both the source and target languages, sharing it between the encoder and decoder. We used the Transformer base model \cite{t2t_vaswani:et:al}. We trained the model on the Google Cloud Platform (8 vCPUs, 30 GB RAM) and followed the standard ``Noam'' learning rate decay,\footurl{https://nvidia.github.io/OpenSeq2Seq/html/api-docs/optimizers.html} see \citet{attention:vaswani:et:al} or \citet{transformer_tip:popel:et:al} for more details. Our starting learning rate was 0.2, and we used 8000 warm-up steps. The model was trained using 200k training steps and 3k validation steps, and the checkpoints were saved at 3k steps.

\begin{table}[t]
    \centering
    \footnotesize
    \begin{tabular}{lrrr}
    \toprule
        \multirow{2}{*}{\textbf{Set}} & \multirow{2}{*}{\textbf{Sentences}} & \multicolumn{2}{c}{\textbf{Tokens}}\\ \cmidrule{3-4}
        & & English  & Hausa  \\
        \midrule
        Train   & 9,632   &35,634 &32,142 \\
        Dev     & 1,204   &4,508 &4,112 \\
        Test    & 1,204   &4,554 &4,084 \\
        \midrule
        Total   & 12,040  & 44,696 & 40,338 \\
        \bottomrule
    \end{tabular}
    \caption{Statistics of our data used in the  English$\leftrightarrow$Hausa text-only and multimodal translation.}
    \label{tab:train_test}
\end{table}

\paragraph{Fine-tuning}
We also employed fine-tuning a large pre-trained model on our domain. This approach has been shown to leverage monolingual data and multilingualism to build a better translation model \cite{adelani-etal-2022-thousand}. We used the M2M-100 pretrained encoder-decoder model \citep{fan2020beyond}. The model was built to translate between 100 language pairs, including Hausa and 16 other African languages. Specifically, we fine-tuned the 418 million parameter version of M2M,\footurl{https://huggingface.co/facebook/m2m100_418M} for three epochs. We used a maximum of 128 tokens for both the target and source sentences and a beam size of 5 during decoding. We trained the model on Google Colab (1 GPU, Tesla T4).


\begin{figure*}[!t]
    \centering
    \includegraphics[clip, trim=0cm 0cm 4.1cm 0cm, width=12cm]{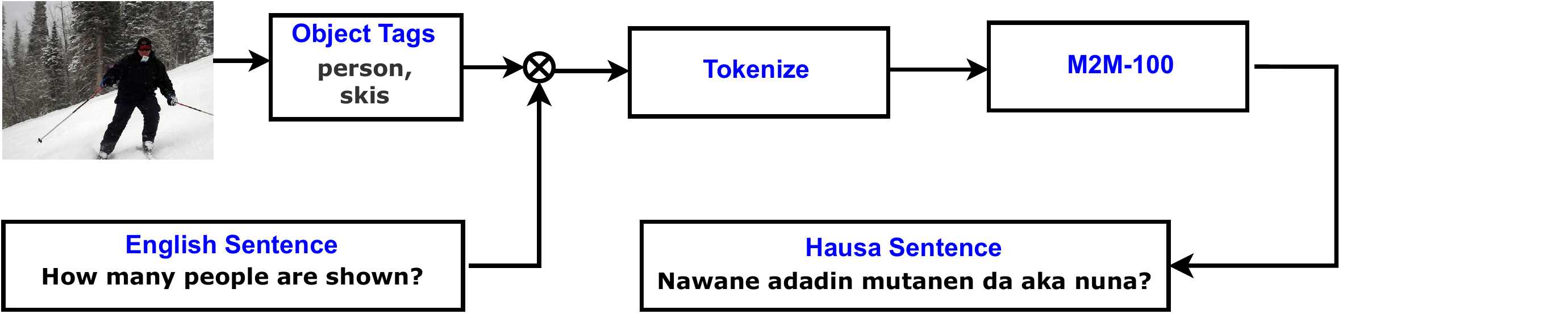}
    \caption{Multimodal machine translation. The object tags are extracted from images and the English source text input to the M2M-100 to generate the Hausa translation output.
    }
    \label{fig:mmt_hausa}
\end{figure*}

\begin{figure*}[ht]
    \centering
    \includegraphics[height=5cm]{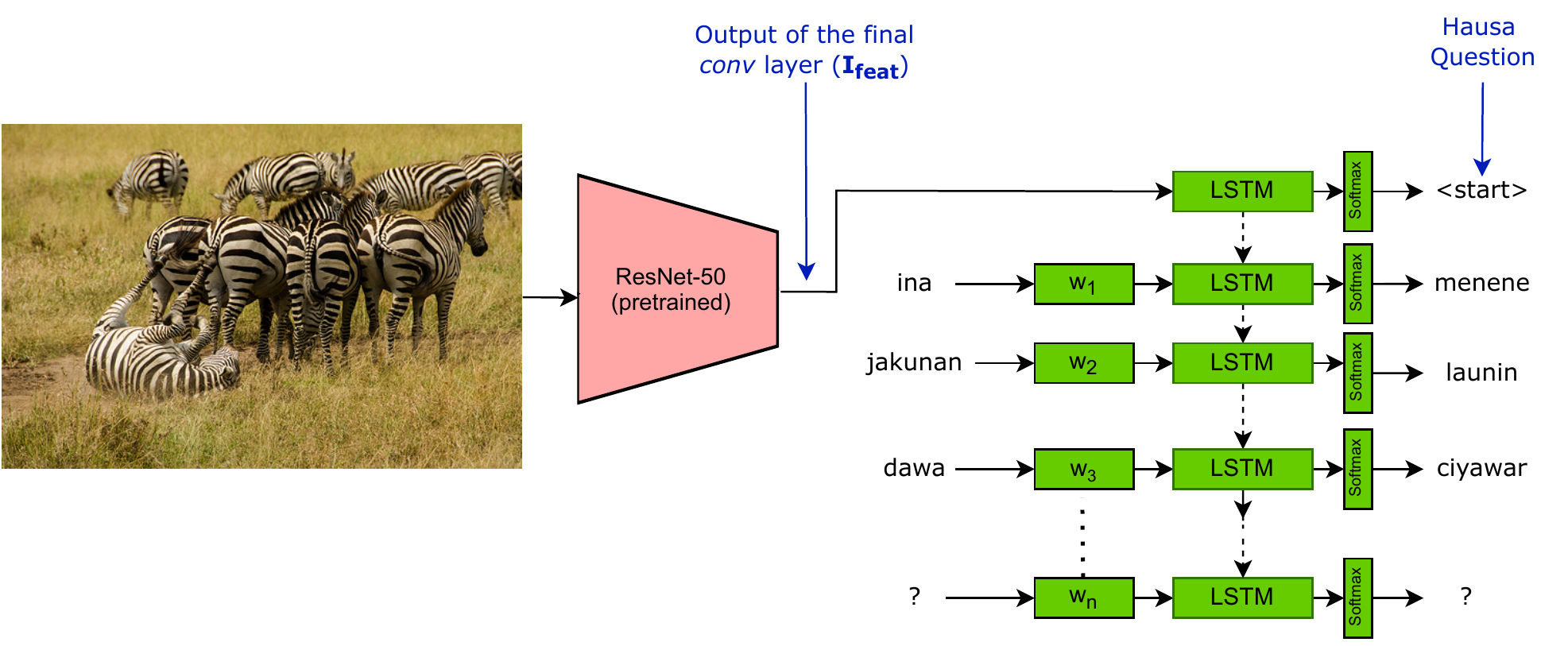}
    \caption{Architecture of Visual Question Elicitation using ResNet-50 and LSTM. The training question was \ ``\textbf{ina jakunan dawa?}'' (\textbf{gloss:} where are the zebras?). When the image was passed during inferencing, the question elicited was ``\textbf{menene launin ciyawar?}'' (\textbf{gloss:} what color is the grass?) }
    \label{fig:ic_hausa}
\end{figure*}

\subsubsection{Multi-Modal Translation}
\label{sect_mt_multitext}

We used the QA pairs of English and Hausa and the associated images for multimodal machine translation (MMT). Multimodal translation involves utilizing the image modality and the English text for translation to Hausa. It extracts automatically learned features from the image to improve translation quality. We take the MMT approach using object tags derived from the image \cite{parida-etal-2021-multimodal}.

We first extract the list of English object tags for a given image using the pre-trained Faster RCNN \cite{Ren_Faster_R_CNN} with ResNet101 \cite{he2016deep} backbone. We consider up to top 10 object tags for each image based on their confidence scores.
    The object tags are concatenated to the English sentence, which needs to be translated into Hausa. The concatenation uses the special token `\#\#’ as the delimiter, followed by comma-separated object tags. Adding object labels enables the otherwise text-only model to utilize visual concepts which may not be readily available in the original sentence, and to supply context information for easier disambiguation. The English sentences and object tags are fed to the encoder of a text-to-text Transformer model, as shown in \Cref{fig:mmt_hausa}.

We used the pre-trained M2M-100 Transformer. We trained the model on the Google Cloud Platform (1 GPU, NVIDIA T4). For comparison, we keep the dataset division the same as the text-only translation, as shown in \Cref{tab:train_test}. 



\subsection{Visual Question Elicitation}
\label{sect_ic}
Similar to image captioning, we used the images and associated questions to train an automatic visual question elicitation (VQE) model.
We extracted visual features using the images and fed them to an LSTM decoder. The decoder generates the tokens of the caption autoregressively using a greedy search approach \cite{soh2016learning}. Trained to minimize the cross-entropy loss on the questions from the training data \cite{yu2019multimodal} was minimized. The architecture is illustrated in \Cref{fig:ic_hausa}.


\paragraph{Image encoder}
All the images were resized to 224$\times$224 pixels, and features from the whole image were extracted to train the model. The feature vector is the output of the final convolutional layer of ResNet-50. It is a 2048-dimensional feature representation of the image. The encoder module is a fixed feature extractor and, thus, non-trainable.

\paragraph{LSTM decoder}
A single-layer LSTM, with a hidden size of 256, was used as a decoder. The dropout is set to 0.3. During training, for the LSTM decoder, the cross-entropy loss is minimized and computed using the output logits and the tokens in the gold caption. Weights are optimized using the Adam optimizer \cite{Adam} with an initial learning rate of 0.001. Training is halted when the validation loss does not improve for ten  epochs. We trained the model for 100 epochs.

\begin{table}[t]
    \centering
    \footnotesize
    \begin{tabular}{lrrr}
    \toprule
        \multirow{2}{*}{\textbf{Set}} & \multirow{2}{*}{\textbf{Sentences}} & \multicolumn{2}{c}{\textbf{Tokens}}\\ \cmidrule{3-4}
        & & English  & Hausa  \\
        \midrule
        Train   & 4816   &27,091 &23,148 \\
        Dev     & 602   &3,306 &2,876 \\
        Test    & 602   &3,320 &2,798 \\
        \midrule
        Total   & 6,020  &33,717 &28,822 \\
        \bottomrule
    \end{tabular}
    \caption{Statistics of our data (questions) used in the \textbf{\textit{Visual Question Elicitation}} task: the number of sentences and tokens.}
    \label{tab:train_test_ie}
\end{table}

\paragraph{VQE Dataset}
We used the Hausa Visual Genome \cite{abdulmumin-etal-2022-hausa} and HaVQA datasets to build our Hausa vocabulary, resulting in 7679 Hausa word types for question generation.
The question elicitation experiment was carried out using the 1,555 images present in the HaVQA dataset. For training and evaluation of the visual question elicitation, we have considered images and questions as shown in \Cref{tab:train_test_ie}. As in VQE, we only considered images and their associated questions ignoring answers which are not necessary for the question generation, the statistics of the dataset differ from the multimodal translation tasks.  


\section{Results and Discussion}
\label{sect_results}
This section presents the results obtained after implementing the experiments described in \Cref{sect_application}.

\subsection{Visual Question Answering}
We employed state-of-the-art language and vision models for Visual Question Answering to report a viable baseline. \Cref{tab:vqa_result} presents the different image encoders we use to obtain our experiment results in combination with the Hausa BERT-based text encoder. The text encoder remains the same across all these experiments. The \textit{WuPalmer} \cite{wu1994verb}
score was chosen as the metric to evaluate the baselines.
The WuPalmer metric measures semantic similarity between words
based on their depth in a lexical hierarchy and the depth of their common ancestor. The metric ranges from 0 to 1, with higher values indicating greater similarity. It is widely employed in tasks such as word sense disambiguation, information retrieval, and semantic relatedness estimation.

From the results reported, the Data-Efficient Image Transformers (DeiT, \citealp{touvron2021training} 
model proposed by Facebook yielded the best results in our architecture. It reached a score of 30.85 and became our best-performing baseline. ViT-base and BEiT Large yielded scores of 
28.90 and 27.75, respectively. ViT Large reported a WuPalmer score of 29.67.
The DeiT models utilize a distillation token to transfer knowledge from a teacher to a student model through backpropagation. This transfer occurs via the self-attention layer, involving the class token (representing the global image representation) and patch tokens (representing local image features). The distillation token interacts with these tokens, assimilating important information from the teacher model and effectively transferring its knowledge. As a result, the student model trained with the distillation token demonstrates improved performance compared to models trained solely with supervised learning.

\begin{table}[t]
    \centering
    \resizebox{\columnwidth}{!}{%
        \begin{tabular}{@{}llr@{}}
        \toprule
            Image Encoder & Text Encoder & WuPalmer Score \\ \midrule
            BEiT-large-P-224 & Bert-base-Hausa & 27.76\\
            ViT-base-P-224   & Bert-base-Hausa & 28.91\\
            ViT-large-P-224   & Bert-base-Hausa & 29.67\\
            \textbf{DeiT-base-P-224} & \textbf{Bert-base-Hausa} & \textbf{30.86}\\
        \bottomrule
        \end{tabular}%
    }
    \caption{Results of the proposed baseline for Visual Question Answering on our HaVQA dataset. It uses the Multimodal Transformer-based architecture.}
    
    \label{tab:vqa_result}
\end{table}

\begin{figure}[!htb]
    \centering
    \includegraphics[width=0.8\columnwidth]{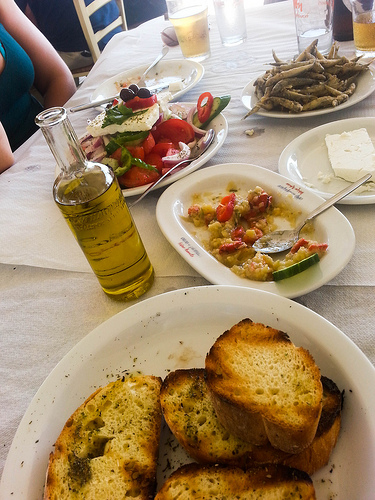}
    \caption{Example of a prediction by the VQA model. The question was ``\textbf{ina cin abincin nan ke faruwa?}'' (\textbf{gloss:} where is this meal taking place?). The ground truth was ``\textbf{a restaurant}'' and the predicted answer was ``\textbf{kan tebur}'' (\textbf{gloss:} on the table)}.
    \label{fig:vqa_example}
\end{figure}

In our study, we conducted manual validation of the results generated by the Visual Question Answering (VQA) model. Our analysis revealed that the model exhibited higher performance when tasked with answering questions that required one-word answers. In these cases, the model consistently provided precise answers for the majority of questions and achieved a very good for the remaining ones.

The training dataset used for training the VQA model consisted of 5500 instances, while the test dataset comprised 520 instances. To provide further insight into the distribution of answers, we presented \Cref{fig:length_precent}, which plots the distribution of word counts in the answers.

By focusing on questions that necessitate one-word answers, our study aimed to explore the extent to which the VQA model can excel in a more restricted task akin to classification. The choice to emphasize single-word answers allowed us to investigate the model's capabilities within a specific context and assess the potential impact of this narrowed scope on its performance.
The observed errors  were mainly associated with cases where there is a dominant object in the picture. The dominant object is returned as the answer regardless the question, see the example prediction in \Cref{fig:vqa_example}. More sample VQA outputs are provided in \Cref{tab:vqa_result_sample} in the Appendix. Example 4 in \Cref{tab:vqa_result_sample} illustrates the same problem when answering the question ``Wacce \textbf{dabba} ce ta fito?'' (\textbf{gloss:} What \textbf{animal} is shown?). Some systems respond with the word ``\textbf{ciyawa}'' (grass), because it is the dominant element in the picture.

\subsection{Machine Translation}
The text-to-text and multi-modal translation model results are shown in \Cref{tab:text-only_trans_results}. For the text-only translation, fine-tuning the Facebook M2M-100 model on the questions and answers for English$\rightarrow$Hausa translation delivers a score by about $+8.4$ BLEU points better than training a Transformer model, and $+11.6$ for Hausa$\rightarrow$English. The multimodal translation model achieved a decent performance comparable to the text-only translation ($-0.8$ BLEU points). 

The better performance by the text-only translation model is expected because, unlike in the Visual Genome dataset \cite{abdulmumin-etal-2022-hausa}, the sentences in HaVQA are mostly unambiguous and, hence, do not require the context that was provided by the images. Also, it is possible that the text captions extracted from the image brought different synonyms than what the single reference translation in Hausa expects. This situation would lead to a comparably good translation quality when assessed by humans but a decreased BLEU score.

\begin{table}[t]
    \centering
    \footnotesize
    \begin{tabular}{lrr}
        \toprule
        \textbf{Method} & 
        \textbf{English}$\rightarrow$\textbf{Hausa} & \textbf{Hausa}$\rightarrow$\textbf{English} \\
        \midrule
        \textbf{Text-Only } & &  \\
        \midrule
        Transformer & 27.1 & 47.1 \\
        M2M-100 & 35.5 & 58.7 \\
        \midrule
        \textbf{MultiModal} & &  \\
         \midrule
         M2M-100  & 26.3 &  - \\
        \bottomrule
    \end{tabular}
    \caption{Results of text-only and multimodal translation on the HaVQA test set.}
    \label{tab:text-only_trans_results}
\end{table}


\subsection{Visual Question Elicitation}

Because it is difficult to measure the quality of the generated questions using automatic evaluation metrics, we manually evaluated the sample-generated questions, relying on a native Hausa speaker. We sampled about 10\% of the elicited questions and subjected them to manual evaluations. We categorized each of the sampled questions as either ``\textbf{Exact}'', ``\textbf{Correct}'', ``\textbf{Nearly Correct}'' and ``\textbf{Wrong}''. We present the distribution of these classes in \Cref{fig:vqe_quality_dist}, and provide some samples in \Cref{tab:vqe_result_sample} in the Appendix.

\begin{figure}[t]
    \centering
    \resizebox{0.8\columnwidth}{!}{%
    \begin{tikzpicture}
        \begin{axis}[
            symbolic x coords={Exact, Correct, Nearly Correct, Wrong},
            xtick=data,
            yticklabel={\pgfmathprintnumber\tick\%},
            width=\columnwidth,
            ylabel=Percentage,
            xlabel=Quality
          ]
            \addplot[ybar,fill=blue!50] coordinates {
                (Exact, 5)
                (Correct, 67)
                (Nearly Correct, 7)
                (Wrong, 22)
            };
        \end{axis}
    \end{tikzpicture}
    }
    \caption{Quality distribution of automatically generated questions.}
    \label{fig:vqe_quality_dist}
\end{figure}
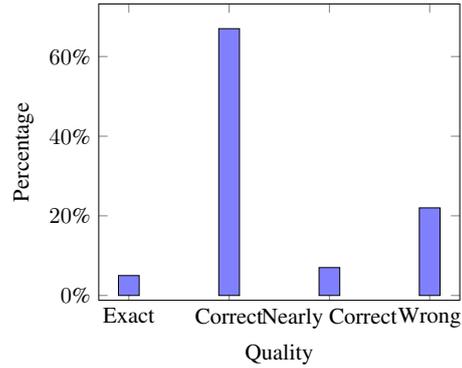

All the generated predictions were valid (reasonable) questions, with all but 3 (99.5\%) having the question mark (`\textbf{$\mathord{?}$}') appended at the end of the question. The distribution of the question types are: ``menene/me/mene'' (what)--64.1\%, ``ina/a ina'' (where)--26.4\%, ``yaushe/da yaushe'' (when)--2.8\%, ``waye/wacce/wana/wanne'' (who)--3.7\%, ``meyasa'' (why)--1.5\%, ``nawane/nawa'' (how much)--1.32\% and ``yaya'' (how)--0.2\%.
 
\section{Conclusion and Future work}
\label{sect_conclusion}

We present HaVQA, a multimodal dataset suitable for many NLP tasks for the Hausa language, including visual question answering, visual question elicitation, text and multimodal machine translation, and other multimodal research.

The dataset is freely available for research and non-commercial usage under a Creative Commons Attribution-NonCommercial-ShareAlike 4.0 License at: \url{http://hdl.handle.net/11234/1-5146}.
We released our experimental code through Github.\footnote{\url{https://github.com/shantipriyap/HausaVQA/tree/main}}

Our planned future work includes: \textit{i)} extending the dataset with more images and QA pairs, \textit{ii)} providing ground truth for all images for image captioning experiments, and \textit{iii)} organizing a shared task using HaVQA. 

\section*{Ethics Statement}
\label{sect_ethics}

We do not envisage any ethical concerns. The dataset does not contain any personal, or personally identifiable, information, the source data is already open source, and there are no risks or harm associated with its usage.

\section*{Limitations}
\label{sect_limitation}

The most important limitation of our work lies in the size of the HaVQA dataset. However, substantial further funding would be needed to resolve this. For the baseline multimodal experiments, we did not use the image directly but resorted to extracting textual tags and including them in the text-only translation input. A tighter fusion technique may give better performance. 

\section*{Acknowledgements}
The HaVQA dataset was created using funding from the HausaNLP research group. This work is supported by Silo AI, Helsinki, Finland. 
This work has received funding from the grant 19-26934X (NEUREM3) of the Czech Science Foundation and has also been supported by the Ministry of Education, Youth, and Sports of the Czech Republic, Project No. LM2018101 LINDAT/CLARIAH-CZ. 




\bibliography{anthology,acl2023}
\bibliographystyle{acl_natbib}

\appendix

\section*{Appendix}

\section{Annotators and Validators Recruitment}
\label{sect_annotator_recruitment}
We recruited Hausa natives from the team of experienced translators at the HausaNLP research group as annotators and validators. The team of annotators consisted of 4 females and 3 males, while the validators included 3 females and 4 males. Each member of the annotation/validation team have at least an undergraduate degree. They all reside in different parts of Northern Nigeria.

\section{Annotation Guidelines}
\label{sect_annotation_guidelines}
The following guidelines were provided to the native Hausa annotators and validators:
\begin{enumerate}
    \item Remember to read the Hausa typing rules. Before starting annotation, test it once and report for any issues.
    \item The annotator must be a native speaker of the Hausa language.
    \item Look at the image before annotating.
    \item Try to understand the task, i.e., translate the questions and answers into the Hausa language.
    \item Do not use any Machine translation system for annotation.
    \item Do not enter dummy entries for testing the interface.
    \item Data will be saved at the backend.
    \item Press the Shift Key on the virtual keyboard for complex consonants.
    \item Contact the coordinator for any clarification/support
\end{enumerate}

\onecolumn
\section{Question-Type Answer Distribution}
\label{sec:ques_type_answer}
\input{answer_quest_dist.tex}

\onecolumn
\section{Visual Question Answer Sample Predictions}
\label{sec:vqa_pred}
\begin{figure*}[!htb]
    \centering
    \begin{tabular}{lp{4cm}lrrcc}
    \toprule
        \multirow{2}{*}{\textbf{S/N}} & \multirow{2}{*}{\textbf{Example}} & \multirow{2}{*}{\textbf{Model}} & \multicolumn{2}{c}{\textbf{Instances}} & \multirow{2}{*}{\textbf{Prediction}} & \multirow{2}{*}{\textbf{Gloss.}}\\
        &&& \textbf{\textit{Test}} & \textbf{\textit{Train}} & &\\
        \midrule
        \textbf{1.} & \multirow{5}{*}{\includegraphics[height=3.8cm]{images/2343297.jpg}} &&&&& \\
        && microsoft/beit-large &520&5500&\textit{rana}&daytime \\
        && google/vit-base &520&5500&\textit{rana}&daytime \\
        && facebook/deit-base &520&5500&\textit{rana}&daytime \\
        && google/vit-large &520&5500&\textit{rana}&daytime \\
        &&&&&& \\
        &&&&&& \\
        &&&&&& \\
        &\multicolumn{2}{l}{\makecell[l]{
        \textbf{Question:} \textit{Yaushe aka \begin{tfour}\m{d}\end{tfour}auki wannan hoton?}\\
        ~\textbf{gloss.} When was this photo taken?}} & 
        \multicolumn{4}{l}{\makecell[l]{
        \textbf{Answer:} \textit{\textit{rana}}\\
        ~\textbf{gloss.} Daytime}}\\ \midrule
        
        \textbf{2.} & \multirow{5}{*}{\includegraphics[height=3.8cm]{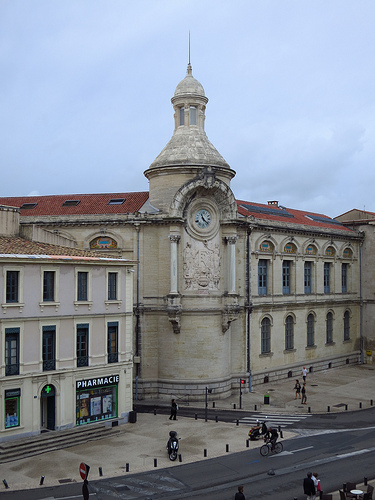}} &&&&& \\
        && microsoft/beit-large &520&5500&\textit{Biyu}&Two \\
        && google/vit-base &520&5500&\textit{Biyu}&Two \\
        && facebook/deit-base &520&5500&\textit{Biyu}&Two \\
        && google/vit-large &520&5500&\textit{Biyu}&Two \\
        &&&&&& \\
        &&&&&& \\
        &&&&&& \\
        &\multicolumn{2}{l}{\makecell[l]{
        \textbf{Question:} \textit{Hawa nawa ne a gefen hagun ginin?}\\
        ~\textbf{gloss.} How many floors is the left of the building?}} & 
        \multicolumn{4}{l}{\makecell[l]{
        \textbf{Answer:} \textit{uku}\\
        ~\textbf{gloss.} three}}\\ \midrule
        
        \textbf{3.} & \multirow{5}{*}{\includegraphics[height=3.8cm]{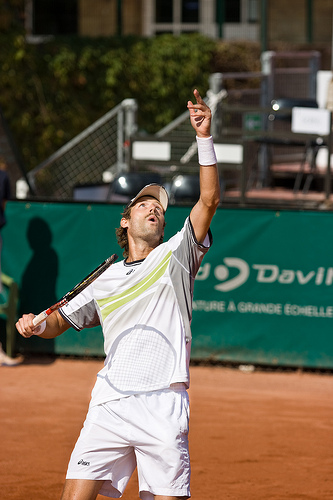}} &&&&& \\
        && microsoft/beit-large &520&5500&\textit{Kore}&Green \\
        && google/vit-base &520&5500&\textit{kore}&green \\
        && facebook/deit-base &520&5500&\textit{kore}&green \\
        && google/vit-large &520&5500&fari&white \\
        &&&&&& \\
        &&&&&& \\
        &&&&&& \\
        &\multicolumn{2}{l}{\makecell[l]{
        \textbf{Question:} \textit{Menene launin rigar sa?}\\
        ~\textbf{gloss.} What color is his shirt?}} & 
        \multicolumn{4}{l}{\makecell[l]{
        \textbf{Answer:} \textit{fari}\\
        ~\textbf{gloss.} white}}\\ \midrule
        
        \textbf{4.} & \multirow{5}{*}{\includegraphics[width=4cm]{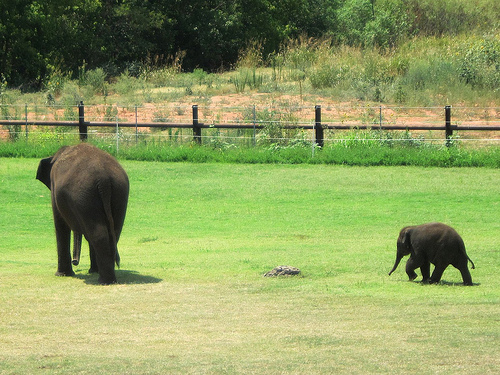}} &&&&& \\
        && microsoft/beit-large &520&5500&\textit{ciyawa}&grass \\
        && google/vit-base &520&5500&\textit{ciyawa}&grass \\
        && facebook/deit-base &520&5500&\textit{dawa}&wild \\
        && google/vit-large &520&5500&\textit{dawa}&wild \\
        &&&&&& \\
        &&&&&& \\
        &&&&&& \\
        &\multicolumn{2}{l}{\makecell[l]{
        \textbf{Question:} \textit{Wacce dabba ce ta fito?}\\
        ~\textbf{gloss.} What animal is shown?}} & 
        \multicolumn{4}{l}{\makecell[l]{
        \textbf{Answer:} \textit{giwa}\\
        ~\textbf{gloss.} elephant}}\\
        \bottomrule
    \end{tabular}
    \caption{Examples of answers generated by the VQA model. All the models were trained using a batch size of 16.}
    \label{tab:vqa_result_sample}
\end{figure*}

\onecolumn
\section{Visual Question Elicitation Sample Predictions}
\label{sec:vqe_pred}
\begin{figure*}[!htb]
    \centering
    \begin{tabular}{p{7.5cm}p{7.5cm}}
    \toprule
        \textbf{Example 1} & \textbf{Example 2} \\
    \midrule
        \textbf{Exact} & \textbf{Correct}\\
        \includegraphics[width=7cm]{images/2382489.jpg} & \includegraphics[height=7cm]{images/2351388.jpg} \\
        \textbf{Ref. Question.}: \textit{Mene launin ciyawar?} & \textbf{Ref. Question.}: \textit{A ina aka \begin{tfour}\m{d}\end{tfour}auki hoton?} \\
        ~~\textbf{Gloss}: What is the color of the grass? & ~~\textbf{Gloss}: Where was the picture taken?\\
        \textbf{Pred. Question.}: \textit{menene launin ciyawar?} & \textbf{Pred. Question.}: \textit{me ra\begin{tfour}\m{k}\end{tfour}umin dawan yakeyi?} \\
        ~~\textbf{Gloss}: what is the color of the grass? & ~~\textbf{Gloss}: what is the giraffe doing? \\
    \midrule
    \midrule
        \textbf{Example 3} & \textbf{Example 4} \\
    \midrule
        \textbf{Nearly Correct} & \textbf{Wrong}\\
        \includegraphics[width=7cm]{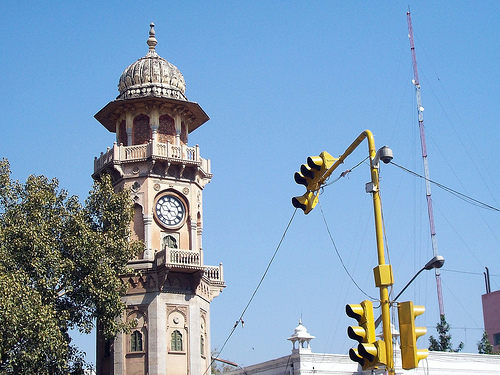} & \includegraphics[width=7cm]{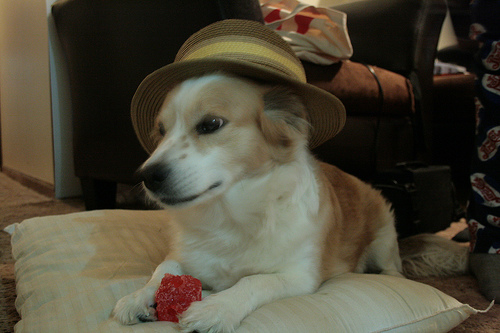} \\
        \textbf{Ref. Question.}: \textit{Ina agogon hasumiya?} & \textbf{Ref. Question.}: \textit{Akan me karen yake?} \\
        ~~\textbf{Gloss}: Where is the tower clock? & ~~\textbf{Gloss}: Where is the dog lying?\\
        \textbf{Pred. Question.}: \textit{ina fitilolin mota?} & \textbf{Pred. Question.}: \textit{menene launin idon magen?} \\
        ~~\textbf{Gloss}: where are the car lights? & ~~\textbf{Gloss}: what is the color of the cat's eye?\\
    
    \bottomrule
    \end{tabular}
    \caption{Examples of Questions elicited by the VQE model.}
    \label{tab:vqe_result_sample}
\end{figure*}

\onecolumn
\section{Glossaries}
\label{sec:gloss}
{\myfontsize\tabcolsep=3pt
    
    \begin{longtable}{llllll}
        
        \toprule
        Hausa & Gloss & Hausa & Gloss & Hausa & Gloss \\
        \cmidrule(lr){1-2}\cmidrule(lr){3-4}\cmidrule(lr){5-6}
        \endfirsthead
    
        \multicolumn{6}{l}{{\tablename} \thetable{} -- Continued} \\ \toprule
        Hausa & Gloss & Hausa & Gloss & Hausa & Gloss \\
        \cmidrule(lr){1-2}\cmidrule(lr){3-4}\cmidrule(lr){5-6}
        \endhead
    
        \midrule
        \multicolumn{6}{l}{Continued on Next Page\ldots} \\
        \bottomrule
        \endfoot
    
        \bottomrule \\
        \caption{Gloss of Hausa words used in \Cref{fig:hau_eng_q_tags,fig:hau_ques_ans_dist}.}
        \label{tab:glossaries}\\
        \endlastfoot
    
        \emph{a} & in & \emph{da yaushe} & when & \emph{kore} & green (s.) \\
        \emph{a benci} & on bench & \emph{daya} & one & \emph{koriya} & green \\
        \emph{a bishiya} & on tree & \emph{\begin{tfour}\m{d}\end{tfour}aya} & one & \emph{kuliya} & cat \\
        \emph{a gona} & on farm & \emph{doki} & horse & \emph{kwa\begin{tfour}\m{d}\end{tfour}o} & frog \\
        \emph{a hagu} & on left & \emph{duhu} & dark & \emph{kwala-kwale} & canoe \\
        \emph{a hanya} & on way (road) & \emph{dutse} & stone & \emph{kwalekwale} & canoe \\
        \emph{a ina} & where (is/are) & \emph{falo} & parlour & \emph{la'asar} & evening \\
        \emph{a jaka} & in bag & \emph{falon} & the parlour & \emph{linzami} & bridle \\
        \emph{akan (me)} & on what & \emph{fanko} & empty & \emph{lu`u-lu`u} & diamond \\
        \emph{a kamara} & on camera & \emph{fara} & white (she) & \emph{mace} & woman \\
        \emph{a kirji} & on chest & \emph{fari} & white (he) & \emph{mage} & cat \\
        \emph{a ruwa} & in water & \emph{fata} & skin & \emph{magen} & the cat \\
        \emph{a sama} & in air & \emph{fika-fikai} & wings & \emph{mai yasa} & why is \\
        \emph{a yaushe} & when & \emph{fili} & field & \emph{mata} & woman \\
        \emph{adadin} & the quantity & \emph{firisbi} & frisbee & \emph{matar} & the woman \\
        \emph{almakashi} & scizzors & \emph{fiza} & pizza & \emph{me} & what \\
        \emph{agogon} & the clock & \emph{fuska} & face & \emph{me aka/ya/ta} & what does \\
        \emph{agwagwa} & duck & \emph{gado} & bed & \emph{me yake da} & what has \\
        \emph{ayaba} & banana & \emph{gajimare} & cloud & \emph{me yasa} & why \\
        \emph{azurfa} & silver & \emph{gilashi} & glass & \emph{mene} & what \\
        \emph{babu} & nothing & \emph{gine-gine} & buildings & \emph{menene} (s.) & what is [it] \\
        \emph{babur} & motorcycle & \emph{gini} & building & \emph{menene} (pl.) & what are \\
        \emph{bacci} & sleep & \emph{giwa} & elephant & \emph{meyasa} & why [did] \\
        \emph{ba\begin{tfour}\m{k}\end{tfour}a} & black (she) & \emph{giwar} & the elephant & \emph{meye} & what \\
        \emph{ba\begin{tfour}\m{k}\end{tfour}i} & black (he) & \emph{guda nawa} & how many & \emph{mutane} & people \\
        \emph{bakowa} & nobody & \emph{gudu} & run & \emph{murabba'i} & square/quarter \\
        \emph{bambaro} & straw & \emph{haske} & light & \emph{mutum} & person \\
        \emph{bango} & wall & \emph{hoton} & the image & \emph{mutumi} & person (m.) \\
        \emph{barci} & sleep & \emph{hudu} & four & \emph{mutumin} & the man \\
        \emph{bas} & bus & \emph{hu\begin{tfour}\m{d}\end{tfour}u} & four & \emph{namiji} & male \\
        \emph{basbal} & baseball & \emph{hula} & cap & \emph{namijin} & the man \\
        \emph{bayyananne} & clear & \emph{ina} & where (is) & \emph{nawa} & how much/mine \\
        \emph{benci} & bench & \emph{inane} & where is & \emph{nawane} & how much [is]/it's mine \\
        \emph{bishiyoyi} & trees & \emph{iyo} & swimming & \emph{na wane} & how much [is]/it's mine \\
        \emph{bishiyu} & trees & \emph{ja} & red & \emph{raktangula} & rectangle \\
        \emph{biyu} & two & \emph{jaririn} & the baby & \emph{rana} & day/sun \\
        \emph{bulo} & block/brick & \emph{kaka} & autumn & \emph{rawaya} & yellow \\
        \emph{ciyawa} & grass & \emph{kamara} & camera & \emph{rufe} & close [it] \\
        \emph{daga (me)} & from what & \emph{kare} & dog & \emph{ruwa} & water \\
        \emph{daga ina} & from where & \emph{karkanda} & rhinoceros & \emph{saniya} & cow \\
        \emph{da'ira} & round & \emph{karnuka} & dogs & \emph{saniyoyi} & cows \\
        \emph{da me} & with what & \emph{\begin{tfour}\m{k}\end{tfour}asa} & sand & \emph{saukowa} & coming down \\
        \emph{dame} & with what & \emph{katako} & wood & \emph{sanwic} & sandwich \\
        \emph{damisa} & tiger & \emph{keke} & bicycle & \emph{shago} & store \\
        \emph{dare} & night & \emph{kofi} & cup & \emph{shinge} & wall \\
        \emph{dawakai} & horses & \emph{koraye} & green (pl.) & \emph{shu\begin{tfour}\m{d}\end{tfour}i} & blue \\
        \emph{su waye} & who are they & \emph{wa yake da} & who has & \emph{wata lamba} & what number \\
        \emph{suwaye} & who are they & \emph{wacce} & what/which is (she) & \emph{waya} & phone/who did ... \\
        \emph{suya} & frying & \emph{wace} & what/which (she) & \emph{wayake} & who is ... (he) \\
        \emph{ta ina} & where & \emph{wace dabba} & what animal & \emph{waye} & who is it (he) \\
        \emph{ta yaya} & how & \emph{waje} & outside & \emph{wucewa} & walking past \\
        \emph{tabarau} & glass & \emph{wake} & who is ... (he/she) & \emph{ya} & how/sister \\
        \emph{tafiya} & walking & \emph{wana} & what/which is (it) & \emph{yaya} & how/sister \\
        \emph{tangaran} & ceramic/china & \emph{wane} & what/which/who is (he) & \emph{yamma} & evening \\
        \emph{tanis} & tennis & \emph{wanene} & who is it (he) & \emph{yarinyar} & the girl \\
        \emph{tashi} & wake/stand & \emph{wani} & what/which (he) & \emph{yaro} & boy \\
        \emph{tasi} & taxi & \emph{wani abu} & what material & \emph{yaron} & the boy \\
        \emph{tauraro} & star & \emph{wani iri} & what type & \emph{yaushe} & when \\
        \emph{tayel} & tie/tile & \emph{wani iri/yanayi} & what kind & \emph{yaushene} & when is it \\
        \emph{titi} & road & \emph{wani kala} & what color & \emph{zagayayye} & round \\
        \emph{tsaro} & security & \emph{wani lokaci} & what time & \emph{zagaye} & round \\
        \emph{tsayuwa} & standing & \emph{wani siffa} & what shape & & \\
        \emph{tunkiya} & sheep & \emph{wani wasa} & what sport &  & \\
        \emph{tunku} & bear & \emph{wanne} & what/which is (he) &  & \\
        \emph{uku} & three & \emph{washe} & clear &  & \\
        \emph{wa} & who & \emph{wasu} & what/which (plural) &  & \\
    
    \end{longtable}

}

\twocolumn




\end{document}